\documentclass[journal]{IEEEtran}
\usepackage{cite}
\usepackage{amsmath,amssymb,amsfonts}
\usepackage{algorithmic}
\usepackage{algorithm}
\usepackage{graphicx}
\usepackage{textcomp}
\usepackage{tabularx}
\usepackage{array}
\usepackage{booktabs}
\usepackage{pdfpages}
\def\BibTeX{{\rm B\kern-.05em{\sc i\kern-.025em b}\kern-.08em
    T\kern-.1667em\lower.7ex\hbox{E}\kern-.125emX}}
\begin{document}
\title{Magnetically Self-Sealed MR Haptic Actuator With PWM-Based Excitation and High-Fidelity Torque Control}

\author{Dong Qiang, Tian Yuan, Song Yang, Kequan Xia, Thomas Reddyhoff, Yikun Zhang, Cheng Cheng,~\IEEEmembership{Member, IEEE}, and Min Yu,~\IEEEmembership{Member, IEEE}%
\thanks{Dong Qiang, Tian Yuan, Song Yang, Kequan Xia, Thomas Reddyhoff, Yikun Zhang, and Min Yu are with the Department of Mechanical Engineering, Imperial College London, London SW7 2AZ, U.K.}%
\thanks{Cheng Cheng is with the School of Artificial Intelligence and Automation, Huazhong University of Science and Technology, Wuhan 430074, China.}%
\thanks{Corresponding author: Min Yu (e-mail: m.yu14@imperial.ac.uk).}%
\thanks{Preprint notice: This work has been submitted to IEEE for possible publication. Copyright may be transferred without notice, after which this version may no longer be accessible.}%
}

\markboth{Preprint Submitted to IEEE/ASME Transactions on Mechatronics}{Qiang \MakeLowercase{\textit{et al.}}: Magnetically Self-Sealed MR Haptic Actuator With PWM-Based Excitation and High-Fidelity Torque Control}

\maketitle

\begin{abstract}
Accurate and stable torque rendering is essential for safe and perceptive human--machine interaction. Magnetorheological fluid (MRF)-based actuators offer a compact and rapidly controllable solution for haptic feedback, but their practical implementation requires reliable fluid sealing, low-hysteresis excitation, accurate torque control, and stable long-duration operation. This article presents an integrated MRF haptic system featuring a compact magnetically self-sealed rotary actuator, low-hysteresis PWM operation, high-fidelity model-based torque rendering, and stable performance during long-time operation. Magnetostatic simulation guides the arrangement of magnetic and nonmagnetic materials to focus flux in the multidisk torque and permanent-magnet sealing regions, enabling a maximum 600~N\,$\cdot$\,mm/A output. Experiments show that higher PWM frequencies reduce hysteresis. At 10~kHz, the response is represented by a nonlinear model that varies with the direction and speed of torque change. The real-time controller combines feedforward, hysteresis compensation, PI feedback, and sliding-mode correction. Compared with PID, it reduces square-wave overshoot, undershoot, and steady-state RMSE by 77.4\%, 61.9\%, and 68.3\%, respectively. It tracks sinusoidal and biomechanics-model-based references, and a 1.5-h test shows only a 2.5~$^{\circ}$C rise near the coil with no clear tracking loss. This high-fidelity torque rendering will fundamentally transform human–robot collaboration by making interactions safer, more efficient, and more intuitive. 
\end{abstract}


\section{Introduction}
\label{sec:introduction}
\IEEEPARstart{H}{aptic} feedback actuators are key enabling components in human--machine interfaces, where they convert interaction information from the task environment into perceivable force or torque cues for the operator. By restoring physical information that is otherwise attenuated or unavailable, these actuators can improve interaction awareness, manipulation precision, and operational safety in teleoperated and human--machine systems. For example, in robot-assisted surgery, the operator must often infer tool--tissue interaction through limited sensory channels, making accurate force feedback particularly valuable for perceiving contact conditions and guiding delicate manipulation \cite{bergholz2023benefits,patel2022haptic}. However, practical deployment remains challenging because high-fidelity force and torque rendering requires not only accurate control, but also compact actuation, low parasitic impedance, and stable long-duration operation under repeated use \cite{patel2022haptic}.

Conventional haptic interfaces based on motor–gear or transmission-heavy architectures often face inherent tradeoffs among torque capability, transparency, mechanical complexity, and thermal behavior. In surgical and other high-precision teleoperation scenarios, these limitations are further exacerbated by friction, backlash, reflected inertia, and the difficulty of integrating dedicated force-sensing hardware in constrained environments \cite{patel2022haptic,Gao2023}. As a result, even when satisfactory short-term tracking performance can be achieved, maintaining consistent and controllable force output in a compact and practically deployable system remains a nontrivial challenge.

Magnetorheological fluids (MRFs) provide an attractive alternative for haptic actuation because their rheological state can be rapidly and reversibly modulated by an external magnetic field, enabling compact torque generation and tunable resistive output \cite{KANG2018649,pisetskiy2021high,Choi2016StateOfTheArt}. Compared with transmission-heavy actuation routes, MRF-based interfaces are well suited to direct-drive or low-transmission configurations, which are beneficial for reducing reflected inertia and improving transparency \cite{pisetskiy2021high}. Moreover, the field-dependent input–output relationship of MRF devices offers a natural basis for control-oriented modeling and torque regulation \cite{KANG2018649,pisetskiy2021high,Choi2016StateOfTheArt}. These advantages have motivated a number of MRF-based haptic devices and medical interface prototypes \cite{najmaei2016design,yang2010development,heo2020knob,yang2021realwalk}. Nevertheless, existing studies have mostly focused on proof-of-concept actuation capability or short-duration control performance, while several system-level issues remain insufficiently addressed for practical deployment.

In particular, an MRF haptic actuator intended for high-fidelity torque rendering must simultaneously satisfy several requirements that are strongly coupled but often studied separately: 1) a compact structure with a stable operating region and reliable fluid sealing, 2) low thermal drift and repeatable output under long-duration operation, 3) a sufficiently reproducible drive condition for control-oriented characterization, and 4) a closed-loop control architecture capable of handling nonlinear backbone behavior, hysteresis, and residual uncertainty without excessive implementation complexity. These issues are especially important under PWM-driven operation, because the drive strategy itself can affect the apparent hysteresis, controllability, and repeatability of the actuator. Therefore, the main challenge is not only to design an MRF actuator, but to establish an integrated MRF haptic actuator system in which structural design, drive condition, open-loop characterization, and real-time control are jointly considered.

To address these coupled requirements, this article presents an integrated MRF haptic system featuring a compact magnetically self-sealed rotary actuator, low-hysteresis PWM operation, high-fidelity model-based torque rendering, and stable performance during sustained operation. The main contributions are summarised as follows:

\begin{enumerate}
    \item A compact magnetically self-sealed MRF rotary haptic actuator is developed to integrate a central coil-driven multidisk torque-generation region with permanent-magnet-assisted sealing regions. Guided by magnetostatic simulation, magnetic and nonmagnetic materials are arranged to concentrate the magnetic flux in the intended working and sealing regions, enabling controllable torque output and magnetic self-sealing without continuous electrical excitation.

    \item A systematic study demonstrates that increasing the PWM carrier frequency reduces torque hysteresis and improves response repeatability, identifying an optimum 10-kHz operating regime and establishing a control-oriented representation comprising a nonlinear backbone and residual hysteresis that depends on the direction and speed of torque change.

    \item A real-time model-based controller combining feedforward, hysteresis compensation, PI feedback, and boundary-layer sliding-mode correction is proposed for accurate torque rendering during both rapid changes and steady operation.

    \item Application-oriented experimental validation demonstrates that the proposed system achieves high-fidelity torque rendering under representative and biomechanics-model-based references, maintains thermal stability, and supports 1.5~h of continuous operation without evident tracking degradation.
\end{enumerate}

\section{Magnetostatic Design and Functional Validation of the Magnetically Self-Sealed MRF Haptic Actuator}

The proposed actuator was designed to meet three coupled requirements of haptic actuation: compact and controllable torque generation, reliable MRF sealing, and efficient utilization of electrical excitation. To this end, the central coil-driven working region and the peripheral permanent-magnet-assisted sealing regions are configured as functionally separated magnetic paths, directing the available coil excitation primarily toward torque generation while maintaining the sealing field without additional continuous coil excitation. This section introduces the haptic requirements and torque rendering principle, details the structural and magnetic functional zoning, and evaluates the resulting field distribution and self-sealing condition through magnetostatic simulation and Mason-number analysis.

\subsection{Application Scenario and Torque Rendering Principle}

\begin{figure}[!t]
\centerline{\includegraphics[width=\columnwidth]{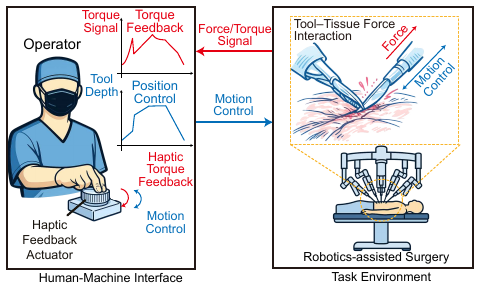}}
\caption{Representative application example of the proposed MRF rotary haptic actuator in robot-assisted surgery. The operator manipulates the user-side haptic interface to generate motion commands for the surgical robot, which reproduces the commanded motion in the task environment and performs tool--tissue interaction. The resulting interaction force/torque signal is transmitted back to the user-side interface and rendered by the MRF actuator as controllable resistive torque, thereby providing the operator with real-time haptic feedback during manipulation.}
\label{fig0}
\end{figure}

The proposed device is developed as an MRF rotary haptic actuator for controllable torque transmission and force-feedback rendering. During operation, the actuator generates resistive torque in response to the operator's manipulation, so that interaction forces from the task environment can be rendered back to the user. This function is achieved through the field-dependent rheological behavior of magnetorheological fluid. Under no or weak magnetic excitation, the MRF behaves close to a viscous fluid and allows low-resistance rotation. When a magnetic field is applied, the suspended particles form chain-like structures along the field direction, increasing the apparent yield stress and shear resistance. The transmitted torque between the stator and rotor can therefore be continuously regulated by adjusting the magnetic excitation.

Fig.~\ref{fig0} illustrates the role of the proposed actuator in a generic haptic-rendering framework, with surgical teleoperation shown as a representative example. The operator manipulates the user-side interface, while interaction force in the task environment converted into a torque reference. The MRF haptic actuator then renders this reference as controllable resistive torque at the user side, thereby closing the force-feedback loop between the operator and the interaction environment.

Supplementary Material~S1 places the target torque range of the proposed actuator in relation to first-source human perception data, including low-force hand perception, manual and upper-limb force discrimination, and joint torque perception. Force-based data are converted to an equivalent rotary-torque scale using a 50 mm effective moment arm, while JND values are retained as relative perceptual thresholds. Under a 15~V PWM input, the proposed actuator reaches approximately 300~N\,$\cdot$\,mm and provides an adjustable range of approximately 20--300~N\,$\cdot$\,mm, corresponding to a controllable torque span of about 280~N\,$\cdot$\,mm based on Section~\ref{sec:oploop}. This places the output range within the order of magnitude of reported human kinesthetic perception levels and supports the selected torque range.

\subsection{Structural Design and Functional Zoning}

\begin{figure}[!t]
\centerline{\includegraphics[width=\columnwidth]{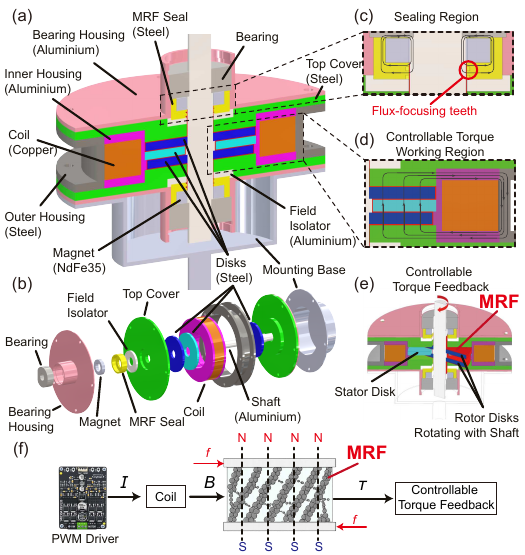}}
\caption{MRF haptic actuator architecture and operating principle of the proposed magnetically self-sealed MRF haptic actuator. (a) Overall sectional view of the integrated rotary actuator. (b) Exploded view of the prototype assembly. (c) Enlarged sealing region, where the permanent magnet forms a dedicated sealing flux path and the flux-focusing teeth locally enhance the magnetic field for MRF sealing. (d) Enlarged working region enabling efficient field-dependent modulation of the magnetorheological effect for controllable torque generation. (e) Torque-feedback mechanism of the rotary haptic interface. (f) PWM-controlled MRF shear mechanism, in which the drive current excites the coil, the resulting magnetic field induces particle-chain structures across the shear gap, and the resistance of these structures to relative disk motion produces a continuously adjustable resistive torque. The sealing and working regions are designed as magnetically decoupled functional zones, so that the sealing field stabilizes the MRF distribution without significantly disturbing the controllable torque-generation region.}
\label{fig1}
\end{figure}

The actuator was filled with MRHCCS4-C magnetorheological fluid (Liquids Research Limited, Batch MR523). To realize torque rendering in a compact rotary module, the proposed actuator integrates controllable MRF torque generation and magnetic fluid sealing within a single architecture. As shown in Fig.~\ref{fig1}, the MRF haptic actuator is divided into a central working region for field-dependent torque generation and peripheral sealing regions for magnetic fluid sealing and operating-state stabilization. In the working region, the stator--rotor disk configuration forms multiple MRF shear gaps, where the applied magnetic field modulates the MRF state and determines the transmitted torque. The magnetic circuit is designed so that the dominant field is approximately normal to the disk shear surfaces, promoting field-induced particle-chain formation across the gaps and increasing the controllable shear resistance.

A key feature of the design is the magnetic self-sealing structure integrated into the same rotary module. Permanent magnets provide local bias fields in the sealing regions, while flux-focusing teeth intensify the field near the sealing interfaces where fluid retention is most critical. Therefore, the sealing region functions as an actively designed magnetic sealing zone rather than a passive mechanical boundary, helping suppress MRF migration and improve output repeatability during rotary operation. The working and sealing regions are supported by magnetically decoupled functional flux paths, allowing the sealing field to be locally enhanced without significantly disturbing the torque-generation field. Nonmagnetic components are also used around the active MRF regions to reduce parasitic flux leakage and preserve the intended field concentration. Through this design, controllable torque generation, magnetic self-sealing, and magnetic-circuit coordination are achieved within one compact haptic actuator. The key design parameters are summarized in Table~\ref{tab:key_design_parameters}.

The operating principle of the actuator can therefore be summarized as follows. During operation, the coil-generated magnetic field modulates the MRF state in the working region, producing a controllable resistive torque between the stator and rotor disks for haptic feedback. At the same time, the permanent-magnet-assisted sealing regions confine the MRF and stabilize its spatial distribution during repeated rotary motion.

\begin{table}[t]
\caption{Key design parameters of the proposed MRF haptic actuator}
\label{tab:key_design_parameters}
\setlength{\tabcolsep}{4pt}
\begin{tabular}{|p{42pt}|p{128pt}|p{60pt}|}
\hline
Symbol &
Parameter &
Value \\
\hline
$N_c$ &
Coil turns &
1000 \\
\hline
$D_s$ &
Main shaft diameter &
6 mm \\
\hline
$D_j$ &
Actuator outer diameter &
68.4 mm \\
\hline
$H_j$ &
Actuator height &
19.8 mm \\
\hline
$g_w$ &
MRF working gap &
0.2 mm \\
\hline
$g_s$ &
Seal gap &
0.1 mm \\
\hline
$L$ &
Characteristic sealing length &
2 mm \\
\hline
$R_w$ &
Effective working radius &
15 mm \\
\hline
$R_s$ &
Characteristic sealing radius &
3 mm \\
\hline
$t_d$ &
Disk thickness &
2 mm \\
\hline
$n_g$ &
Number of active shear gaps &
4 \\
\hline
\end{tabular}
\end{table}

\subsection{Magnetic Self-Sealing Assessment Based on Mason-Number Analysis}
\label{subsec:mason_main}

\begin{figure}[!t]
\centering
\includegraphics[width=0.7\columnwidth]{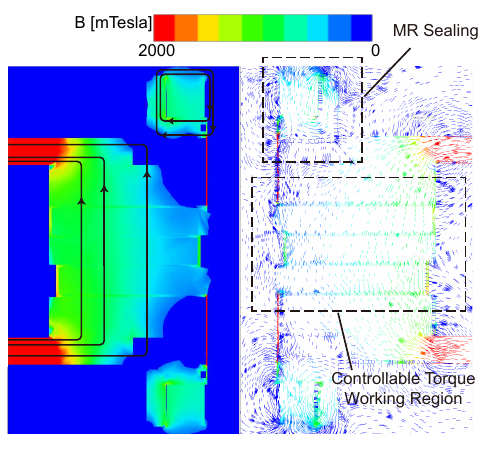}
\caption{Magnetostatic field distribution of the proposed MRF haptic actuator under a coil current of 500~mA. The magnetic flux is concentrated primarily in the disk-based working region and the peripheral sealing regions, supporting controllable torque generation and magnetically assisted fluid sealing, respectively.}
\label{fig2}
\end{figure}

Following the magnetostatic verification of the sealing-region field (Fig.~\ref{fig2}), the sealing condition was assessed using the Poiseuille--Couette Mason-number framework of Liang \textit{et al.}~\cite{liang2018dynamic}, together with related studies on magnetically assisted MR-fluid sealing~\cite{Kordonski1996MRFSeal,Zhang2018IEEEFerrofluidMRFSeal,Hegger2019SmartSealing,Kubik2019SMSShaftSeal}. The framework compares the pressure-driven load, wall-driven shear, and field-dependent yield resistance within the thin sealing gap. For the proposed MRF haptic actuator, it was applied under a conservative gravity-loaded condition in which the total contained MRF weight was assumed to act on the lower annular sealing interface. The complete formulation, parameter derivation, applicability assumptions, and uncertainty analysis are provided in Supplementary Material~S3.

Using a representative sealing-gap flux density of approximately $0.2~\mathrm{T}$, a mechanical rotation frequency of $2~\mathrm{Hz}$, and the geometric parameters listed in Table~I, the calculated pressure- and rotation-related Mason numbers were $Mn(p)=0.0626$ and $Mn(\Omega)=185.4$, respectively. The corresponding transition bounds were $Mn(p)_{R1}=0.0108$ and $Mn(p)_{R2}=2.152$. The condition
\begin{equation}
Mn(p)_{R1}<Mn(p)<Mn(p)_{R2}
\end{equation}
therefore places the sealing flow in the \emph{two-region mode}. This classification was retained under one-at-a-time variations of $\pm20\%$ in the yield stress, sealing-gap width, and characteristic leakage-path length, as well as across the combined uncertainty cases evaluated in Supplementary Material~S3. In the most conservative combined case, the lower-bound margin remained $Mn(p)/Mn(p)_{R1}\approx3.09$.

In the two-region mode, the Poiseuille--Couette model predicts a magnetically stabilized plug region attached to the sealing-gap wall, with shear localized in the adjacent yielded region~\cite{liang2018dynamic,Hegger2019SmartSealing,Kubik2019SMSShaftSeal}. This configuration increases the resistance to pressure-driven MRF migration during rotary motion and supports the intended sealing function of the permanent-magnet-assisted sealing region. No visible MRF leakage was observed during the subsequent characterization and controller-tuning experiments. Together, the analytical classification and this qualitative observation provide preliminary support for fluid sealing under the tested conditions and establish the basis for the actuator characterization and torque-rendering studies in Sections~III--V.

\section{Open-Loop Analysis and Control-Oriented Modeling}
\label{sec:oploop}

Following the structural and sealing validation, this section characterises the PWM-driven torque behaviour of the proposed MRF haptic actuator and establishes its control-oriented representation. The effects of PWM carrier frequency, input frequency, and operating range are evaluated to identify a reproducible operating regime and to separate the dominant nonlinear backbone from the residual branch- and rate-dependent hysteresis.

\begin{figure}[!t]
\centerline{\includegraphics[width=\columnwidth]{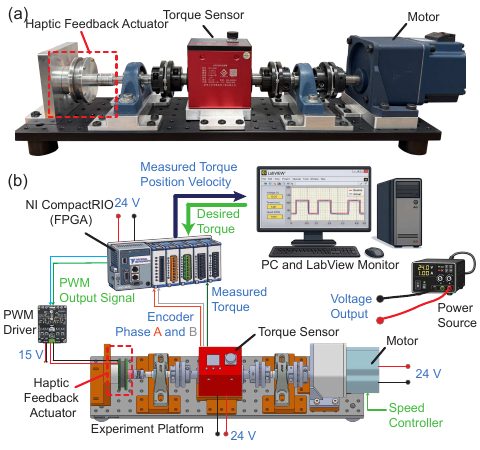}}
\caption{(a) Experimental setup for open-loop characterisation and closed-loop torque-rendering experiments. (b) Signal flow for characterisation and control of the proposed MRF haptic actuator. The test rig integrates motor-driven rotary excitation, torque and encoder measurements, PWM-based power driving, and FPGA--RT real-time implementation.}
\label{fig3}
\end{figure}

\subsection{Experimental Setup and FPGA-Based Data Acquisition}

The open-loop characterization experiments were conducted on the dedicated platform shown in Fig.~\ref{fig3}. The platform integrates the proposed MRF haptic actuator, a drive motor with a speed controller, a DYN-200 torque sensor, an incremental encoder, a PWM driver, an NI CompactRIO-based real-time controller, and a host computer for monitoring and parameter adjustment. The actuator was coupled to the motor through the transmission shaft, and the transmitted torque was measured by the DYN-200 sensor mounted in the driveline. The encoder integrated with the torque sensor, with a resolution of 8000 counts per revolution, measured shaft motion through quadrature phase-$A$ and phase-$B$ signals, enabling synchronized torque and motion measurement under different excitation conditions.

The real-time system was built on an NI cRIO-9035 controller equipped with one NI~9239 analog input module and two NI~9401 digital I/O modules. The NI~9239 acquired the analog torque signal from the DYN-200 sensor; one NI~9401 module acquired the encoder phase-$A$ and phase-$B$ signals, while the other generated the PWM signal. A Cytron MDD10A Rev.~2.0 motor driver received the FPGA-generated PWM command and regulated the electrical excitation applied to the actuator. The motor and torque sensor were supplied at 24~V, while the PWM driver was powered independently at 15~V. This configuration allowed the PWM carrier frequency and duty cycle to be systematically adjusted to examine their influence on torque generation, hysteresis, and output repeatability.

Experiment execution and data acquisition were organized through a hierarchical FPGA--RT architecture. The FPGA layer operated with a 40~MHz onboard clock and handled torque acquisition, encoder counting, PWM generation, and synchronized data logging. The RT layer was executed at 1000~Hz for supervisory control, parameter management, and experiment coordination. This deterministic architecture ensured synchronized actuation, sensing, and logging for long-duration and repeatable open-loop analysis, providing the basis for subsequent control-oriented modeling.

\subsection{Selection of the PWM Operating Frequency}

PWM excitation was adopted to regulate the electrical excitation applied to the MRF actuator in an efficient switch-mode manner. Compared with linear driving, PWM driving reduces driver-stage power dissipation and enables the effective coil excitation to be adjusted through the duty cycle. However, the carrier frequency can directly affect the apparent torque-output behavior. Low-frequency PWM may introduce current ripple and discontinuous excitation, whereas excessively high frequency may increase switching losses without providing further response improvement. Therefore, the PWM carrier frequency was selected by considering both drive efficiency and torque-output repeatability.

\begin{figure}[!t]
\centerline{\includegraphics[width=\columnwidth]{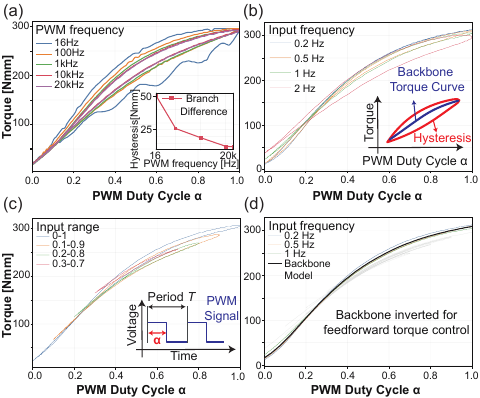}}
\caption{Open-loop characterization of the proposed MRF haptic actuator.
(a) PWM-frequency-dependent torque--input hysteresis loops and extracted hysteresis metrics.
(b) Hysteresis dependence on input frequency under the selected 10 kHz PWM regime.
(c) Hysteresis dependence on operating range under the selected 10 kHz PWM regime.
(d) Feature extraction and control-oriented model construction under the selected 10 kHz PWM regime.}
\label{fig4}
\end{figure}

The influence of PWM carrier frequency was experimentally evaluated using a sinusoidal duty-cycle input with $f=1~\mathrm{Hz}$ and $\alpha=0\text{--}1$. As shown in Fig.~\ref{fig4}(a), the torque--input hysteresis loop is strongly affected by the carrier frequency. At 16~Hz, the loop is the widest, indicating pronounced branch-dependent response and poor output reproducibility. Increasing the carrier frequency to 100~Hz reduces the loop width, while the remaining hysteresis is still larger than that observed in the kHz range. Once the frequency enters the kHz regime, the loop becomes substantially more compact and the input--output relation becomes more repeatable. The extracted metrics show the same trend: both the branch gap and hysteresis area decrease markedly from 16~Hz to 1~kHz, while only minor changes are observed from 1~kHz to 20~kHz.

These results indicate that increasing the PWM frequency is effective for reducing hysteresis at low carrier frequencies, but the improvement gradually converges in the kHz range. Accordingly, 10~kHz was selected as the nominal operating frequency for subsequent open-loop characterization, control-oriented model construction, and closed-loop torque control, because it lies within the converged low-hysteresis regime without requiring unnecessarily higher switching frequencies.

\subsection{Dynamic Hysteresis and Backbone Extraction at 10~kHz}

After selecting 10~kHz as the nominal PWM operating frequency, further open-loop characterization was conducted to examine the residual nonlinear and hysteretic behavior under this drive condition. The purpose was to determine which part of the torque--input response could be treated as a reproducible backbone and which part should be compensated as branch- and rate-dependent residual hysteresis.

Sinusoidal duty-cycle commands with different input frequencies and operating ranges were applied while keeping the PWM carrier frequency fixed at 10~kHz. Fig.~\ref{fig4}(b) compares the responses under different input frequencies, while Fig.~\ref{fig4}(c) compares the responses under different operating ranges. As shown in Fig.~\ref{fig4}(b), non-negligible residual hysteresis remains even in the selected high-frequency PWM regime. The branch positions do not vary as a simple uniform widening of the loop, indicating that the remaining deviation is both branch- and rate-dependent. Therefore, a purely static inverse map is insufficient for high-fidelity torque rendering, and a residual hysteresis compensation layer is required.

Fig.~\ref{fig4}(c) further shows that the hysteresis behavior depends on the operating range. Compared with narrower mid-range excitation, the full-range input produces a more distorted loop, especially near the two ends of the input domain. This suggests that the middle operating region provides more favorable conditions for stable characterization, while the end regions are more sensitive to amplified hysteresis and local slope variation.

Based on these observations, Fig.~\ref{fig4}(d) extracts a nominal backbone curve from representative broad-range responses under the selected 10~kHz regime. This backbone represents the dominant nonlinear torque--input relation and is used to construct the inverse feedforward map. The residual difference between the measured branches and the backbone is then interpreted as a branch- and rate-dependent hysteretic component. This decomposition preserves the main experimental structure while avoiding independent fitting of each dynamic loop, making it suitable for real-time implementation in the layered closed-loop controller.

\section{Hierarchical Closed-Loop Control Design}

\begin{figure*}[!t]
\centerline{\includegraphics[width=0.819\textwidth]{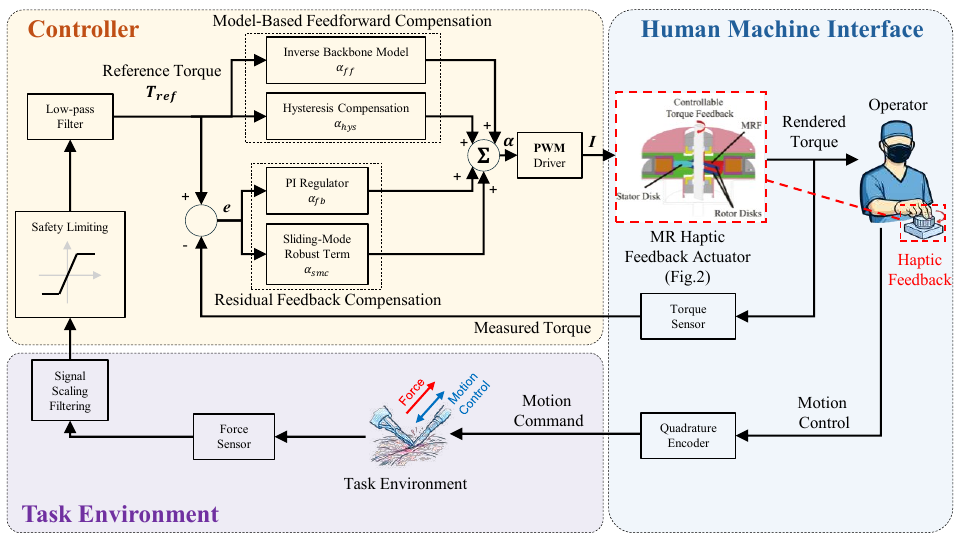}}
\caption{System architecture of the proposed MRF rotary haptic actuator for torque rendering in robot-assisted surgery. The operator commands the surgical robot through the user-side interface, while the resulting tool--tissue interaction force/torque is converted into a torque reference and tracked by the hierarchical controller. The MRF actuator renders the corresponding resistive torque to the operator, thereby closing the real-time haptic feedback loop.}
\label{fig5}
\end{figure*}

Real-time torque rendering with the proposed MRF actuator is affected by the nonlinear backbone, residual hysteresis, and slow actuation-span variation identified in the open-loop characterization. If feedback control is used alone, the controller must compensate the full nonlinear actuation characteristic through tracking error, which can introduce delay and degrade transient torque rendering. Therefore, as shown in Fig.~\ref{fig5}, the proposed controller is formulated as a layered compensation structure. The inverse-backbone feedforward term compensates the dominant nonlinear relation, the branch/rate-dependent term corrects structured hysteresis, the residual PI block regulates the remaining tracking error, and a quasi-steady boundary-layer sliding-mode term improves robustness during plateau segments.

\subsection{Control Architecture and Adaptive Backbone Feedforward}

The unsaturated duty-ratio command is constructed as
\begin{equation}
\tilde{\alpha}_k
=
\eta_{ff}\alpha_{ff,k}
+
\eta_{hys}\alpha_{hys,k}
+
\alpha_{fb,k}
+
\alpha_{smc,k},
\label{eq:alpha_unsat}
\end{equation}
where $\alpha_{ff,k}$ is the inverse-backbone feedforward term, $\alpha_{hys,k}$ is the branch/rate-dependent hysteresis compensation term, $\alpha_{fb,k}$ is the residual feedback contribution generated by the external PI block, and $\alpha_{smc,k}$ is the quasi-steady robust correction term. The gains $\eta_{ff}$ and $\eta_{hys}$ scale the feedforward and hysteresis-compensation layers.

To avoid direct control action on raw measurements, the reference and measured torque signals are first processed by discrete first-order low-pass filters. The filtered signals are denoted as $T_{ref,f,k}$ and $T_{meas,f,k}$, respectively. The filter time constants for the reference and measured torque channels are $T_{f,ref}$ and $T_{f,meas}$, and their values are listed in Table~\ref{tab:hhc_parameters}.

The first control layer maps the filtered reference torque to a nominal duty-ratio command using the inverse backbone model extracted from the 10~kHz open-loop response. Because the effective actuation span of MRF devices may vary slowly with thermal, rheological, and operation-induced conditions~\cite{Li2021RheologicaActaTemperatureMRF,Kariganaur2022SMSTemperatureMRF,Ashtiani2015JMMMPreparationStabilization,Thiagarajan2021AEMPerformanceStability}, a bounded scalar correction factor $g_{ff,k}$ is introduced. The torque-domain input to the inverse map is defined as
\begin{equation}
T_{ff,in,k}
=
T_{min}
+
\frac{T_{ref,f,k}-T_{min}}{g_{ff,k}},
\label{eq:Tff_in}
\end{equation}
where $T_{min}$ is the lower bound of the nominal backbone. When $g_{ff,k}<1$, the plant is interpreted as weaker than the nominal model and a larger feedforward command is generated; when $g_{ff,k}>1$, the command is reduced.

The feedforward command is obtained from the piecewise inverse backbone map. Let the inverse look-up-table breakpoints be denoted by $\{(T_i,\alpha_i)\}_{i=0}^{N}$, with $\alpha_i\in[0,1]$. For
\begin{equation}
T_{ff,in,k}\in[T_i,T_{i+1}],
\end{equation}
linear interpolation gives
\begin{equation}
\alpha_{ff,k}
=
\alpha_i
+
\frac{T_{ff,in,k}-T_i}{T_{i+1}-T_i}
\left(\alpha_{i+1}-\alpha_i\right).
\label{eq:alpha_ff_piecewise}
\end{equation}
This representation preserves the experimentally identified backbone while remaining lightweight for real-time implementation.

\subsection{Branch- and Rate-Dependent Hysteresis Compensation and Residual Regulation}

Although the inverse backbone compensates the dominant nonlinear relation, the open-loop results show that the residual hysteresis depends on both branch direction and excitation rate. The filtered reference increment is
\begin{equation}
\Delta T_{ref,f,k}=T_{ref,f,k}-T_{ref,f,k-1}.
\label{eq:dTref}
\end{equation}
Using a deadband $\varepsilon_T$ to avoid branch chattering near flat reference segments, the branch indicator is updated as
\begin{equation}
b_k=
\begin{cases}
+1, & \Delta T_{ref,f,k}>\varepsilon_T,\\
-1, & \Delta T_{ref,f,k}<-\varepsilon_T,\\
b_{k-1}, & |\Delta T_{ref,f,k}|\le\varepsilon_T .
\end{cases}
\label{eq:branch_state}
\end{equation}
The rate proxy used for compensation scheduling is
\begin{equation}
r_k=
\left|
\frac{\Delta T_{ref,f,k}}{T_s}
\right|.
\label{eq:rate_proxy}
\end{equation}
The hysteresis compensation term is written as
\begin{equation}
\alpha_{hys,k}
=
b_k G_h(\alpha_{ff,k},r_k),
\label{eq:alpha_hys}
\end{equation}
where $G_h(\cdot)$ is constructed from experimentally extracted compensation curves. For $r_k\in[r_j,r_{j+1}]$, the scheduled compensation function is obtained by
\begin{equation}
G_h(\alpha_{ff,k},r_k)
=
(1-\mu_k)h_j(\alpha_{ff,k})
+
\mu_k h_{j+1}(\alpha_{ff,k}),
\label{eq:rate_interp}
\end{equation}
with
\begin{equation}
\mu_k=
\frac{r_k-r_j}{r_{j+1}-r_j}.
\label{eq:mu}
\end{equation}
Here, $h_j(\alpha)$ and $h_{j+1}(\alpha)$ denote neighboring compensation curves identified from representative dynamic hysteresis loops. This formulation treats the residual hysteresis as a structured deviation around the backbone, rather than fitting each full hysteresis loop independently.

After the model-based layers are applied, the remaining tracking error is
\begin{equation}
e_k=T_{ref,f,k}-T_{meas,f,k}.
\label{eq:error}
\end{equation}
The residual feedback term $\alpha_{fb,k}$ is generated by an external PI block acting on $e_k$, while the Formula Node computes the feedforward, hysteresis-compensation, span-adaptation, and robust-correction terms.

To improve robustness against residual uncertainty, slow drift, and operation-induced disturbances, a boundary-layer sliding-mode correction is activated only during quasi-steady reference segments:
\begin{equation}
\chi_{smc,k}
=
\begin{cases}
1, & |\Delta T_{ref,f,k}|\le \kappa_T\varepsilon_T,\\
0, & |\Delta T_{ref,f,k}|> \kappa_T\varepsilon_T,
\end{cases}
\label{eq:smc_gate}
\end{equation}
where $\kappa_T$ relaxes the branch-update deadband for quasi-steady detection. The discrete sliding variable is
\begin{equation}
\sigma_k
=
\lambda_s e_k
+
\eta_{\Delta e}(e_k-e_{k-1}),
\label{eq:sigma}
\end{equation}
and the robust correction is
\begin{equation}
\alpha_{smc,k}
=
\chi_{smc,k} K_s
\operatorname{sat}
\left(
\frac{\sigma_k}{\phi}
\right),
\label{eq:alpha_smc}
\end{equation}
where $K_s$ is the switching gain, $\phi$ is the boundary-layer thickness, and $\operatorname{sat}(\cdot)$ is the saturation function. The quasi-steady gate prevents the robust term from reacting to sharp reference transitions, thereby avoiding excessive command correction at square-wave edges.

\subsection{Slow Span Adaptation and Real-Time Command Limiting}

The scalar span factor $g_{ff,k}$ is updated only under high-platform quasi-steady conditions, so that the learning process is not affected by transient hysteresis or low-torque local deviations. When $T_{ref,f,k}$ is sufficiently above $T_{min}$ and $|\Delta T_{ref,f,k}|$ is small, an instantaneous gain estimate is defined as
\begin{equation}
g_{inst,k}
=
\frac{T_{meas,f,k}-T_{min}}{T_{ref,f,k}-T_{min}}.
\label{eq:g_inst}
\end{equation}
The update law is
\begin{equation}
g_{ff,k+1}
=
\operatorname{sat}_{[g_{min},g_{max}]}
\left[
g_{ff,k}
+
\chi_{g,k}\beta_g
\left(
\bar{g}_{inst,k}-g_{ff,k}
\right)
\right],
\label{eq:gff_update}
\end{equation}
where $\chi_{g,k}$ is the learning gate, $\beta_g$ is the adaptation gain, and $\bar{g}_{inst,k}$ denotes the clipped instantaneous estimate within $[g_{min},g_{max}]$. This bounded slow adaptation reduces the need for manual recalibration when the effective actuation span changes.

Finally, the unsaturated command in \eqref{eq:alpha_unsat} is subjected to a slew-rate limit:
\begin{equation}
\alpha_{rl,k}
=
\alpha_{k-1}
+
\operatorname{sat}_{[-\dot{\alpha}_{max}T_s,\,\dot{\alpha}_{max}T_s]}
\left(
\tilde{\alpha}_k-\alpha_{k-1}
\right),
\label{eq:alpha_rate_limit}
\end{equation}
and is then bounded within the admissible duty-ratio interval:
\begin{equation}
\alpha_k
=
\operatorname{sat}_{[\alpha_{min},\alpha_{max}]}
\left(
\alpha_{rl,k}
\right).
\label{eq:alpha_sat}
\end{equation}
The final command $\alpha_k$ is sent to the PWM driver. The real-time execution sequence is summarized in Algorithm~\ref{alg:hhc}, and the implementation parameters used in the closed-loop experiments are listed in Table~\ref{tab:hhc_parameters}.

\begin{algorithm}[!t]
\caption{Real-Time Hierarchical Hysteresis-Compensated Control}
\label{alg:hhc}
\begin{algorithmic}[1]
\REQUIRE $T_{ref,k}$, $T_{meas,k}$, previous states $T_{ref,f,k-1}$, $T_{meas,f,k-1}$, $b_{k-1}$, $g_{ff,k}$, $e_{k-1}$, $\alpha_{k-1}$
\ENSURE PWM duty-ratio command $\alpha_k$

\STATE Apply first-order low-pass filters to $T_{ref,k}$ and $T_{meas,k}$ to obtain $T_{ref,f,k}$ and $T_{meas,f,k}$.
\STATE Compute $T_{ff,in,k}=T_{min}+(T_{ref,f,k}-T_{min})/g_{ff,k}$
\STATE Obtain $\alpha_{ff,k}$ from the inverse backbone LUT using \eqref{eq:alpha_ff_piecewise}
\STATE Compute $\Delta T_{ref,f,k}=T_{ref,f,k}-T_{ref,f,k-1}$
\STATE Update branch indicator $b_k$ using \eqref{eq:branch_state}
\STATE Compute rate proxy $r_k=|\Delta T_{ref,f,k}/T_s|$
\STATE Compute $\alpha_{hys,k}=b_kG_h(\alpha_{ff,k},r_k)$
\STATE Compute filtered tracking error $e_k=T_{ref,f,k}-T_{meas,f,k}$
\STATE Obtain residual feedback term $\alpha_{fb,k}$ from the external PI block

\IF{$|\Delta T_{ref,f,k}|\le \kappa_T\varepsilon_T$}
    \STATE Compute $\sigma_k=\lambda_s e_k+\eta_{\Delta e}(e_k-e_{k-1})$
    \STATE Compute $\alpha_{smc,k}=K_s\operatorname{sat}(\sigma_k/\phi)$
\ELSE
    \STATE Set $\alpha_{smc,k}=0$
\ENDIF

\STATE Construct $\tilde{\alpha}_k=\eta_{ff}\alpha_{ff,k}+\eta_{hys}\alpha_{hys,k}+\alpha_{fb,k}+\alpha_{smc,k}$
\STATE Apply slew-rate limit to obtain $\alpha_{rl,k}$
\STATE Saturate $\alpha_{rl,k}$ into $[\alpha_{min},\alpha_{max}]$ to obtain $\alpha_k$

\IF{$T_{ref,f,k}>T_H$ and $|\Delta T_{ref,f,k}|\le \varepsilon_g$}
    \STATE Estimate $g_{inst,k}=(T_{meas,f,k}-T_{min})/(T_{ref,f,k}-T_{min})$
    \STATE Update $g_{ff,k+1}$ using \eqref{eq:gff_update}
\ELSE
    \STATE Set $g_{ff,k+1}=g_{ff,k}$
\ENDIF

\STATE Update stored states for the next sampling step
\RETURN $\alpha_k$
\end{algorithmic}
\end{algorithm}

\begin{table}[!t]
\caption{Implementation Parameters of the Proposed HHC Controller}
\label{tab:hhc_parameters}
\centering
\footnotesize
\setlength{\tabcolsep}{2.5pt}
\begin{tabular}{p{45pt}p{58pt}p{110pt}}
\hline
Parameter & Value & Note \\
\hline
$T_s$ 
& $1.0\times10^{-3}~\mathrm{s}$ 
& Real-time sampling period \\

$T_{f,ref}$ 
& $2.0\times10^{-3}~\mathrm{s}$ 
& Reference-torque filter \\

$T_{f,meas}$ 
& $2.0\times10^{-3}~\mathrm{s}$ 
& Measured-torque filter \\

$\eta_{ff}$ 
& $0.66$ 
& Feedforward scaling \\

$\eta_{hys}$ 
& $1.60$ 
& Hysteresis-compensation scaling \\

$T_{min}$ 
& $8.25~\mathrm{N\,mm}$ 
& Lower backbone bound \\

$g_{ff,0}$ 
& $1.00$ 
& Initial span factor \\

$g_{min},\,g_{max}$ 
& $0.95,\;1.05$ 
& Span-factor bounds \\

$\beta_g$ 
& $1.0\times10^{-3}$ 
& Span-adaptation gain \\

$T_H$ 
& $70~\mathrm{N\,mm}$ 
& High-platform learning threshold \\

$\varepsilon_g$ 
& $0.05~\mathrm{N\,mm}$ 
& Flat-reference learning threshold \\

$\varepsilon_T$ 
& $2~\mathrm{N\,mm}$ 
& Branch-update deadband \\

$r_j$ 
& $\begin{array}{@{}c@{}}
92.31,\;461.55,\\
923.10~\mathrm{N\,mm/s}
\end{array}$ 
& Rate breakpoints \\

$K_c$ 
& $0.176$ 
& Residual PI proportional gain \\

$T_i$ 
& $0.18~\mathrm{s}$ 
& Residual PI integral time \\

$K_s$ 
& $2.0\times10^{-3}$ 
& Robust-term gain \\

$\lambda_s$ 
& $5$ 
& Sliding-variable error weight \\

$\eta_{\Delta e}$ 
& $1$ 
& Error-increment weight \\

$\phi$ 
& $5$ 
& Boundary-layer thickness \\

$\kappa_T$ 
& $3$ 
& Quasi-steady gate factor \\

$\alpha_{min},\,\alpha_{max}$ 
& $0,\;1$ 
& Duty-ratio saturation bounds \\
\hline
\end{tabular}
\end{table}

\section{Closed-Loop Experimental Validation}

Closed-loop experiments were conducted to evaluate the torque-rendering performance of the proposed hierarchical hysteresis-compensated control (HHC) strategy. Three controllers were compared under the same experimental conditions: a tuned PID controller, a feedforward--PI controller (FF--PI), and the proposed HHC controller. The PID controller serves as a feedback-only baseline. The FF--PI controller introduces the identified inverse backbone map but does not include the branch/rate-dependent hysteresis compensation or robust correction. The proposed HHC further incorporates the hysteresis compensation and quasi-steady robust correction layers. This comparison therefore evaluates the controller in an ablation-like manner, from pure feedback regulation to backbone compensation and finally to the full hierarchical compensation structure.

The validation references were selected to examine different aspects of torque-rendering performance. Square-wave references were used to assess transient response, overshoot, undershoot, and plateau regulation. A sinusoidal reference was used to evaluate continuous dynamic tracking, including amplitude and phase consistency. Finally, a biomechanics-inspired torque profile was used to examine the ability of the controller to render a task-relevant interaction signal containing both abrupt and slowly varying features.

\begin{figure}[!t]
\centerline{\includegraphics[width=\columnwidth]{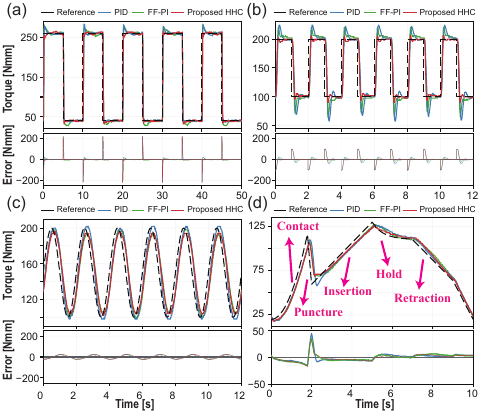}}
\caption{Closed-loop comparison of PID, FF--PI, and the proposed controller under representative torque-rendering tasks. 
(a) Large-amplitude square-wave response. 
(b) Repeated square-wave response. 
(c) Sinusoidal response. 
(d) Biomechanics-inspired interaction torque profile constructed from typical needle--tissue interaction phases~\cite{okamura2004force,mahvash2010mechanics,vanGerwen2012needle}.}
\label{fig:closed_loop}
\end{figure}

\begin{table*}[!t]
\caption{Quantitative Closed-Loop Tracking Performance of Different Controllers}
\label{tab:closed_loop_performance}
\centering
\footnotesize
\setlength{\tabcolsep}{4pt}
\begin{tabular}{p{100pt}p{118pt}c c c c c}
\hline
Reference input & Metric & PID & FF--PI & Proposed HHC & Reduction vs. PID & Reduction vs. FF--PI \\
\hline
Square-wave references$^{\dagger}$
& RMSE [N\,mm] & 34.69 & 32.54 & \textbf{31.32} & 9.7\% & 3.7\% \\
& Avg. response time [s] & 0.102 & 0.082 & \textbf{0.080} & 21.9\% & 2.5\% \\
& Avg. overshoot [N\,mm] & 22.35 & 19.75 & \textbf{5.05} & 77.4\% & 74.4\% \\
& Avg. undershoot [N\,mm] & 19.53 & 21.30 & \textbf{7.43} & 61.9\% & 65.1\% \\
& Steady-state RMSE [N\,mm] & 6.39 & 4.68 & \textbf{2.02} & 68.3\% & 56.8\% \\
\hline
Sinusoidal reference
& Trajectory RMSE [N\,mm] & \textbf{14.43} & 16.59 & 15.95 & -10.5\% & 3.9\% \\
\hline
Biomechanics-inspired profile
& Trajectory RMSE [N\,mm] & 14.61 & 13.15 & \textbf{12.32} & 15.7\% & 6.3\% \\
& Event-window RMSE [N\,mm] $^{\dagger}$ & 49.81 & 40.73 & \textbf{38.29} & 23.1\% & 6.0\% \\
& Hold RMSE [N\,mm] & 9.74 & 6.55 & \textbf{4.83} & 50.4\% & 26.3\% \\
\hline
\end{tabular}

\begin{minipage}{0.96\textwidth}
\footnotesize
$^{\dagger}$All square-wave metrics are averaged over the large-amplitude and repeated square-wave tests; the response time is further averaged over rising and falling transitions. 
For the biomechanics-inspired profile, the event-window RMSE is computed within a short window around the puncture-related transient event to quantify the tracking accuracy during the most abrupt interaction phase.
\end{minipage}
\end{table*}

\subsection{Comparative Tracking Performance Under Canonical References}

Fig.~\ref{fig:closed_loop}(a)--(c) compares the closed-loop tracking responses under canonical square-wave and sinusoidal references. These inputs were used to evaluate complementary torque-rendering characteristics, including transient response, plateau recovery, repeated switching consistency, and continuous dynamic tracking. In the square-wave tests, the abrupt reference transitions induce sharp transient errors, where the PID controller regulates the torque mainly after the tracking error appears and the FF--PI controller reduces the feedback burden through inverse-backbone compensation. However, residual branch-dependent mismatch remains visible near the rising and falling transitions. By further introducing branch/rate-dependent hysteresis compensation and quasi-steady robust correction, the proposed HHC controller provides a more balanced response between transition directions and improves the consistency across repeated cycles. In the sinusoidal test, all controllers follow the smooth reference trend, while the remaining deviations near the peak and valley regions reflect the influence of torque--input nonlinearity and residual hysteresis. Owing to the smooth and continuous variation of the sinusoidal reference, all three controllers achieve comparably satisfactory tracking, and the differences among them are less pronounced than those observed under abrupt square-wave transitions. Overall, the proposed HHC controller achieves the best combined performance across the canonical reference tests, particularly in suppressing transient overshoot and undershoot while maintaining stable continuous tracking. The corresponding quantitative comparison is summarized in Table~\ref{tab:closed_loop_performance}.

\begin{figure}[!t]
\centerline{\includegraphics[width=\columnwidth]{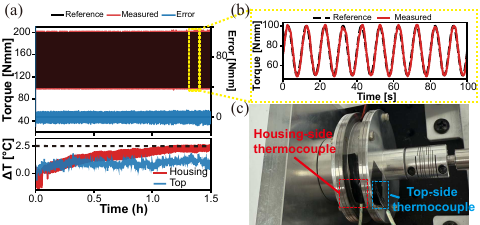}}
\caption{Long-duration closed-loop stability test of the proposed MRF haptic actuator. 
(a) Torque tracking response and temperature variation during continuous operation. 
(b) Enlarged view of the selected tracking segment. 
(c) Thermocouple placement on the housing and top side for temperature monitoring.}
\label{fig:Thermal Stability}
\end{figure}

\begin{figure}[!t]
\centerline{\includegraphics[width=0.9\columnwidth]{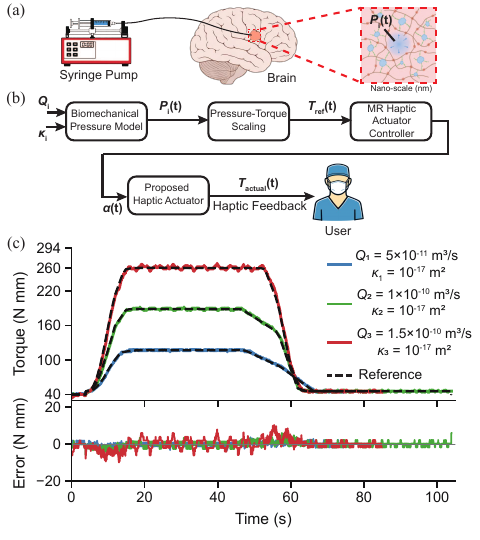}}
\caption{Haptic feedback validation based on \textit{in vivo} validated biomechanical model of brain drug injection. 
(a) Schematic of localized brain drug injection and local pressure generation during injection. 
(b) Signal flow from biomechanical pressure modelling to MRF haptic torque rendering. 
(c) Closed-loop tracking results under three representative flow rate (\(Q_i\)) and tissue permeability (\(\kappa_i\)) conditions, with the corresponding tracking errors shown below.}
\label{fig:biomechanics_haptic_validation}
\end{figure}

\subsection{Long-Duration Torque-Rendering Stability}
\label{subsec:long_duration_stability}

Beyond the short-duration tracking comparisons, the sustained operating performance of the actuator--controller system was evaluated through a 1.5-h continuous test. The thermocouple arrangement and corresponding measurement locations on the haptic actuator are shown in Fig.~\ref{fig:Thermal Stability}(c). As shown in Fig.~\ref{fig:Thermal Stability}(a), the measured torque remained close to the periodic reference throughout the test, while the enlarged segment in Fig.~\ref{fig:Thermal Stability}(b) shows no evident loss of tracking quality during the later stage of operation. The temperature rise at the housing location closest to the coil gradually approached approximately $2.5~^{\circ}\mathrm{C}$, while the top-side temperature rise remained lower over the same period. These results show that stable periodic torque rendering was maintained with limited thermal accumulation under the tested sustained-actuation condition.

\subsection{Haptic Validation for Intracerebral Drug Injection}
\label{subsec:biomechanics_validation}

The blood--brain barrier limits the effectiveness of most systemically administered drugs for brain diseases~\cite{yuan2022effect}, making direct intracranial injection a promising method for targeted brain drug delivery. However, clinical translation remains constrained by the need for precise control of both needle insertion and drug infusion, owing to the extreme softness and structural complexity of brain tissue~\cite{yuan2023porosity, yuan2022microstructurally, yuan2023linking}. During insertion, excessive force or an inappropriate trajectory may damage healthy tissue, blood vessels, or critical brain structures. During infusion, insufficient pressure can substantially prolong the procedure, whereas excessive pressure may damage the tissue, open the interface between the needle and the surrounding tissue, and induce backflow along the needle track~\cite{jamal2022insights}. In this context, this newly developed closed-loop haptic control system could be particularly valuable by enabling clinicians to perceive the mechanical resistance encountered during needle insertion and changes in injection pressure during infusion, thereby supporting safer needle placement and more controlled drug delivery~\cite{yuan2025infusion}. 

Evaluating such a system requires interaction profiles that capture the distinct mechanical events occurring throughout the procedure. The biomechanics-inspired interaction profile in Fig.~\ref{fig:closed_loop}(d) therefore extends the preceding canonical tests by combining abrupt local changes with slowly varying segments associated with needle contact, puncture, insertion, hold, and retraction phases~\cite{yuan2024comprehensive}. 
Following evaluation of this prescribed insertion profile, model-informed rendering during drug infusion was assessed using pressure trajectories generated by a previously developed and \textit{in vivo} validated intracerebral drug injection model, as illustrated in Fig.~\ref{fig:biomechanics_haptic_validation}. This bottom-up, multiscale model links the prescribed infusion flow rate $Q_i$ and tissue permeability $\kappa_i$ to the resulting local pressure response $p_i(t)$, thereby providing a direct biomechanical basis for the haptic reference signal. The framework was derived from fluid tissue interactions at the axonal scale, upscaled to a pressure-dependent tissue permeability formulation, and incorporated into an anatomically realistic whole brain transport model. Its predictions were validated against \textit{ex vivo} tissue injection measurements and \textit{in vivo} sheep brain injection experiments using MRI based tracer tracking~\cite{yuan2025brainInterstitialTransport}. This experimentally validated model therefore enables the rendered haptic feedback to reflect the underlying mechanics of intracerebral infusion, rather than an empirically prescribed waveform.

As shown in Fig.~\ref{fig:biomechanics_haptic_validation}(b), the simulated pressure profile was scaled into a torque reference and supplied to the MRF haptic actuator controller, forming a model-to-haptic signal chain from the biomechanical parameters to the rendered resistive torque. Three cases were evaluated at a fixed permeability of $\kappa=10^{-17}~\mathrm{m}^{2}$, with flow rates of $5\times10^{-11}$, $1\times10^{-10}$, and $1.5\times10^{-10}~\mathrm{m}^{3}/\mathrm{s}$. These conditions produced low-, medium-, and high-amplitude reference profiles, respectively. As shown in Fig.~\ref{fig:biomechanics_haptic_validation}(c), the actuator reproduced the initial pressure rise, plateau-like segment, and subsequent relaxation in all three cases. Together with the canonical and structured-reference results, this experiment demonstrates the feasibility of rendering model-derived, noncanonical torque trajectories over different feedback levels and validates the implemented model-to-haptic signal chain under the tested conditions.

\section{Conclusion} 
This article presented a magnetically self-sealed magnetorheological (MR) rotary haptic actuator for stable and controllable torque rendering. The actuator integrates a central MRF working region with peripheral magnetic sealing regions, enabling torque generation, fluid sealing, and operating-state stabilization within a compact rotary structure. Magnetostatic simulation and Mason-number analysis supported the proposed magnetic self-sealing condition, while open-loop characterization identified a reproducible high-frequency PWM regime for control-oriented modelling. Based on the observed nonlinear backbone and residual hysteresis, a hierarchical hysteresis-compensated control framework was developed using inverse feedforward mapping, branch/rate-dependent hysteresis compensation, PI feedback, and boundary-layer sliding-mode regulation. Comparative experiments showed improved torque-rendering performance, with average square-wave overshoot, undershoot, and steady-state RMSE reduced by 77.4\%, 61.9\%, and 68.3\%, respectively, compared with the PID baseline. Application-oriented tests, including model-derived intracerebral infusion profiles and extended-duration operation, demonstrated accurate torque rendering, sustained stability, and minimal thermal rise.

\section*{Acknowledgment}

This work was supported by Dr. Min Yu's Imperial College Research Fellowship (ICRF)

\bibliographystyle{IEEEtran}
\bibliography{reference}

\begin{IEEEbiography}[{\includegraphics[width=1in,height=1.25in,clip,keepaspectratio]{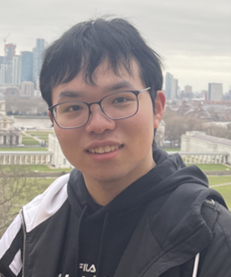}}]{Dong Qiang}
received the M.Eng. degree in mechanical engineering from Imperial College London, London, U.K., in 2025. He is currently pursuing the Ph.D. degree with the Department of Mechanical Engineering, Imperial College London. His research interests include the design and control of magnetorheological-fluid-based haptic interfaces and sensing strategies for monitoring frictional systems.
\end{IEEEbiography}

\begin{IEEEbiography}[{\includegraphics[width=1in,height=1.25in,clip,keepaspectratio]{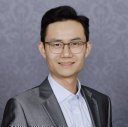}}]{Tian Yuan}
received the Ph.D. degree in mechanical engineering from Imperial College London, London, U.K., in 2023. He is currently a Royal Academy of Engineering Research Fellow with the Department of Mechanical Engineering, Imperial College London. His research interests include multiscale and multiphysics modelling of fluid and mass transport in biological tissues, experimental characterisation of soft-tissue properties, and the optimisation of targeted drug-delivery procedures for the brain and solid tumours.
\end{IEEEbiography}

\begin{IEEEbiography}[{\includegraphics[width=1in,height=1.25in,clip,keepaspectratio]{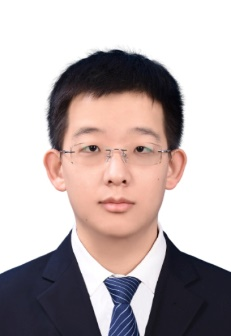}}]{Song Yang}
received the B.S. and M.S. degrees in energy and power engineering from Xi'an Jiaotong University, Xi'an, China, in 2021 and 2024, respectively. He is currently pursuing the Ph.D. degree with the Department of Mechanical Engineering, Imperial College London, London, U.K. His research interests include nondestructive testing techniques for monitoring friction and lubrication processes.
\end{IEEEbiography}

\begin{IEEEbiography}[{\includegraphics[width=1in,height=1.25in,clip,keepaspectratio]{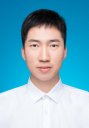}}]{Kequan Xia}
received the Ph.D. degree from Zhejiang University, Hangzhou, China, in 2022. He is currently a Postdoctoral Researcher with Imperial College London, London, U.K. His research interests include triboelectric nanogenerators, self-powered tactile sensing, flexible electronics, and intelligent human--machine interfaces.
\end{IEEEbiography}

\begin{IEEEbiography}[{\includegraphics[width=1in,height=1.25in,clip,keepaspectratio]{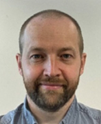}}]{Thomas Reddyhoff}
is an Associate Professor with the Department of Mechanical Engineering, Imperial College London, London, U.K. His research focuses on improving the performance of sliding contacts and frequently involves the development of novel in situ measurement techniques in combination with numerical modelling.
\end{IEEEbiography}

\begin{IEEEbiography}[{\includegraphics[width=1in,height=1.25in,clip,keepaspectratio]{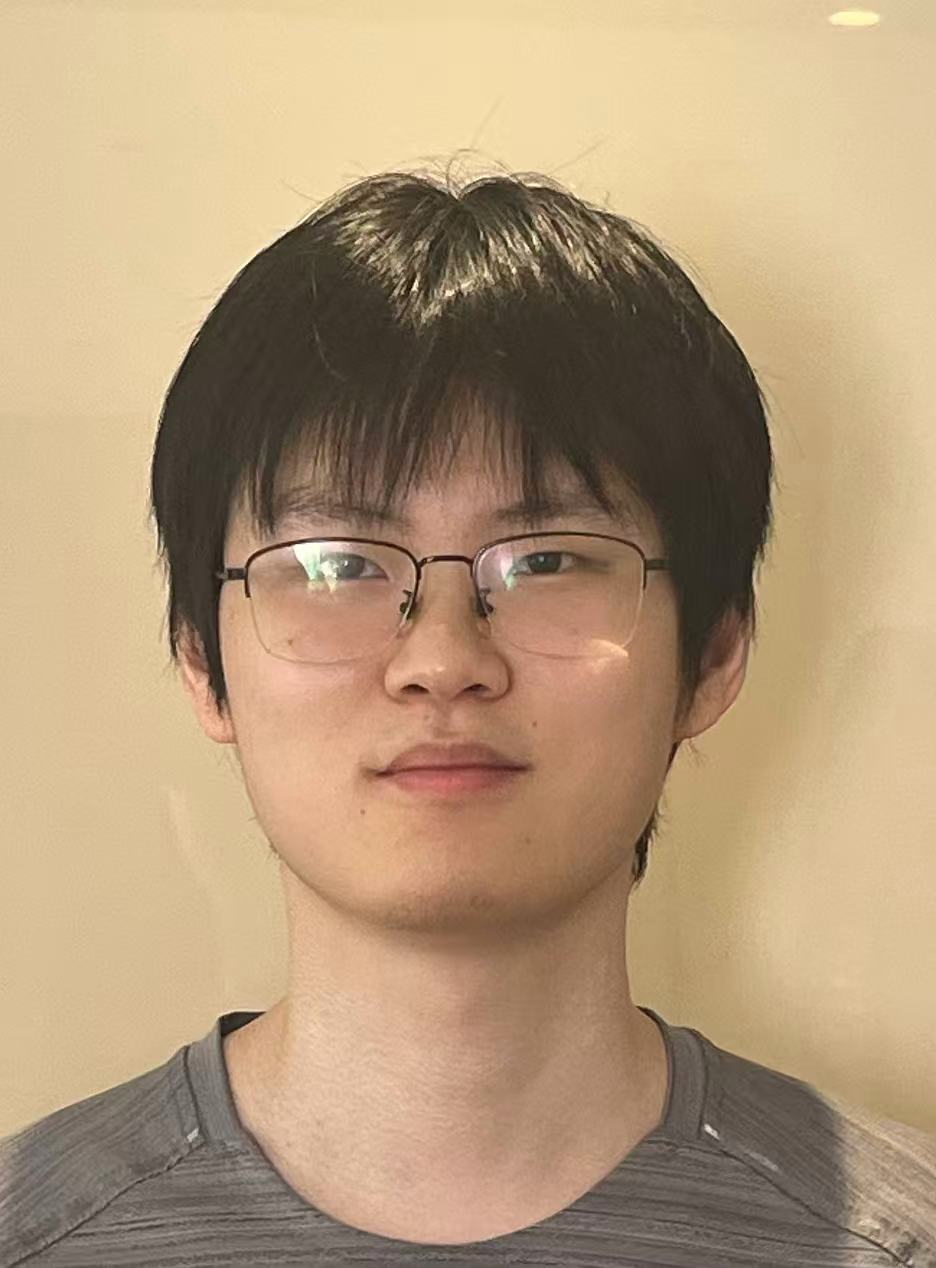}}]{Yikun Zhang}
received the M.Eng. degree in mechanical engineering from Imperial College London, London, U.K., in 2026. His research interests lies in hardware designs of magnetorheological-fluid-based haptic interfaces and steel manufacturing designs for automotive industry.
\end{IEEEbiography}

\begin{IEEEbiography}[{\includegraphics[width=1in,height=1.25in,clip,keepaspectratio]{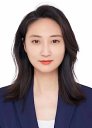}}]{Cheng Cheng}
(Member, IEEE) received the B.Eng. degree in measurement and control technology and instrumentation from Tianjin University, Tianjin, China, in 2012, and the M.Sc. and Ph.D. degrees in control systems from Imperial College London, London, U.K., in 2013 and 2018, respectively. Since 2020, she has been with Huazhong University of Science and Technology, Wuhan, China, where she is currently an Associate Professor with the School of Artificial Intelligence and Automation. Her research interests include robust control, modelling and simulation of mechatronic systems, and deep-learning applications.
\end{IEEEbiography}

\begin{IEEEbiography}[{\includegraphics[width=1in,height=1.25in,clip,keepaspectratio]{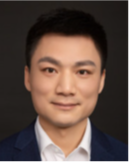}}]{Min Yu}
(Member, IEEE) received the Ph.D. degree in control engineering from Imperial College London, London, U.K., in 2018. He is currently a Research Fellow with the Department of Mechanical Engineering, Imperial College London. His research interests include control and sensing techniques for smart engineering interfaces, mechanical performance enhancement, structural health monitoring, and human--machine interaction.
\end{IEEEbiography}

\clearpage


\section*{Supplementary material}

# Supplementary Material



## Supplementary Material S1: Human perception levels relevant to kinesthetic haptic torque rendering

TABLE S1
HUMAN PERCEPTION LEVELS RELEVANT TO KINESTHETIC HAPTIC TORQUE RENDERING

| Perceptual condition | Reported perceptual level and equivalent rotary-torque scale† |
|---|---|
| Low-force hand perception | Reference forces of 0.1–0.8 N correspond to 5–40 N · mm with a 50 mm effective moment arm. Near-threshold hand-force perception and related low-force discrimination data were reported in [1]–[3]. |
| Manual force discrimination | Reference forces of 2.5–10 N correspond to 125–500 N · mm with a 50 mm effective moment arm. Active thumb–forefinger force JND is approximately 6–8%, with related force-matching data reported in [2]–[4]. |
| Upper-limb and joint torque perception | Active arm-motion Weber fractions of 0.08–0.13 correspond to relative force-change JNDs of approximately 8–13%. Wrist torque-change JND is approximately 12–19%, while elbow torque JND ranges from 97–197 N · mm under static conditions and can reach 799 N · mm under dynamic conditions [3]–[6]. |
| Proposed MRF haptic actuator | At 15 V PWM input, the measured torque range is approximately 20–300 N mm, with a controllable span of about 280 N mm, corresponding to approximately 93% of the maximum output torque based on Section III. This range overlaps representative human kinesthetic perception scales and provides a usable span for torque rendering. |

†Force-to-torque conversions use a 50 mm effective moment arm only to place force-based psychophysical data on the same rotary-torque scale. JND denotes the just-noticeable difference.

Table S1 summarizes representative psychophysical findings relevant to kinesthetic haptic torque rendering and places force-based perception data on a common rotary-torque scale for comparison with the proposed actuator. For force-based studies, an effective moment arm of 50 mm is used solely as a reference conversion and does not imply a specific human–device configuration. The reported data span low-force perception, manual force discrimination, and upper-limb or joint torque discrimination, covering both absolute perceptual levels and relative just-noticeable differences. Within this context, the measured output range of the proposed MRF haptic actuator, approximately 20–300 N · mm with a controllable span of about 280 N · mm, overlaps representative kinesthetic torque scales reported in the literature. This comparison is intended to provide a perception-relevant reference for interpreting the available torque-rendering range rather than to establish device-specific perceptual thresholds, which would require dedicated human-subject experiments.

## Supplementary Material S2: Comparison With Representative MR Haptic Actuators

TABLE S2
COMPARISON WITH REPRESENTATIVE MR-BASED HAPTIC ACTUATORS

| Study | Device / size / mass | Raw torque / excitation | Sealing | Nonlinearity and torque control | Thermal stability test | Engineering validation |
|---|---|---|---|---|---|---|
| Wellborn et al. [7] | Hybrid brake; 38-mm diameter × 14.5 mm; 67.8 g | 462 N mm at 0.5 A | Silicone-bonded mating surfaces | Hysteresis characterized; current regulation | Current selected to limit heating; quantitative test N/R | Torque, hysteresis, step, and frequency tests |
| Chen et al. [8] | Spherical actuator, 3 DoF; 43.8-mm diameter × 33.5 mm; 173.3 g | 341.25 N mm at 1 A | Sealing ring | Hysteresis characterized; reverse-current demagnetization | 8.56-W maximum power; quantitative test N/R | Multiaxis torque and haptic-effect demonstrations |
| Blake and Gurocak [9] | Small and large MR brakes in a wearable glove; 68 / 84 g | 93 / 189 N mm at 0.5 A, before gearing | Stamped silicone seals | Current-based contact rendering | Heating noted above 0.5 A; quantitative test N/R | Six-brake glove and human grasp/stiffness tests |
| Najmaei et al. [10] | Armature MRF clutch; 36-mm diameter × 60 mm; 598 g | 1500 N mm; current N/R | Sealed outer-casing caps | 1-kHz PID with Hall/ANN torque estimation | Heating not significant in reported tests; quantitative data N/R | Teleoperation, needle insertion, and tissue palpation |
| Heo et al. [11] | Rotary haptic knob; envelope and mass N/R | 5.61 N mm at 0.25 A and 90°/s | Sealed and bonded housing cover | Open-loop current–torque mapping | 1.25-W design limit; quantitative test N/R | Torque–current, torque–speed, and frequency tests |
| This work | Self-sealed multidisk rotary actuator; 68.4-mm diameter × 19.8 mm; 325 g | 300 N mm at 0.5 A; 15 V, 10 kHz PWM | Permanent-magnet-assisted self-sealing regions | 1-kHz hysteresis-compensated closed-loop torque regulation | 1.5-h continuous test; ∼2.5°C coil-temperature rise | Canonical and biomechanics-derived torque rendering |

N/R denotes not reported.

Table S2 compares representative MR-based haptic actuators in terms of device configuration, torque output, fluid-confinement approach, nonlinear torque regulation, thermal stability, and engineering validation. Wellborn *et al.* [7] developed a compact hybrid MR brake with high device-level torque and systematically characterized its hysteresis, transient response, and frequency response. Chen *et al.* [8] realized multidirectional torque generation using a compact three-degree-of-freedom spherical MR

actuator. Blake and Gurocak [9] integrated miniature MR brakes into a wearable haptic glove and demonstrated grasping and stiffness rendering. Najmaei *et al.* [10] combined MRF clutches with model-based torque estimation and feedback control for medical teleoperation, needle insertion, and tissue palpation. Heo *et al.* [11] developed a low-power rotary MR knob for compact consumer haptic interfaces. Together, these studies demonstrate the versatility of MR-based devices for compact torque generation and application-specific haptic interaction.

The proposed actuator extends these developments through an integrated rotary architecture that jointly addresses torque generation and magnetic fluid confinement. It provides an adjustable torque range of approximately $20–300\ \mathrm{N\,mm}$ and reaches approximately $300\ \mathrm{N\,mm}$ at a maximum operating current of $0.5$ A under 15-V, 10-kHz PWM excitation. The central multidisk working region provides controllable MR shear resistance, while the peripheral permanent-magnet-assisted regions maintain magnetic fluid confinement without requiring continuous electrical excitation for the sealing function. The sealing mechanism is therefore incorporated directly into the magnetic and mechanical design of the actuator. Its operating condition is further supported by the magnetostatic analysis in the main manuscript and the detailed Mason-number assessment in Supplementary Material S3.

The drive and control strategy are established from the measured behavior of the actuator rather than from a nominal current–torque relationship alone. Open-loop characterization shows that the PWM carrier frequency affects the apparent torque hysteresis and output repeatability. A reproducible 10-kHz PWM regime is consequently selected for control-oriented modeling. Under this operating condition, the measured response is decomposed into a dominant nonlinear backbone and branch- and rate-dependent residual hysteresis. The proposed hierarchical hysteresis-compensated controller combines inverse-backbone feedforward, structured hysteresis compensation, residual PI feedback, and boundary-layer sliding-mode regulation in a 1-kHz real-time control loop.

Among the compared systems, Najmaei *et al.* [10] also demonstrated closed-loop torque regulation using Hall-sensor feedback and an artificial neural network for torque estimation. The present work instead uses direct output-torque measurement and explicitly constructs the controller from the PWM-dependent nonlinear and hysteretic characteristics of the proposed actuator. This establishes a direct progression from drive-condition selection and open-loop characterization to control-oriented modeling and real-time torque regulation.

The system-level validation further examines whether the actuator retains its rendering performance during sustained operation. During the $1.5$-h continuous test, the measured torque remained close to the periodic reference without evident progressive tracking degradation. The temperature rise at the housing location nearest the coil gradually approached approximately $2.5\ ^\circ\mathrm{C}$, while the top-side temperature rise remained lower. The actuator therefore maintained stable periodic torque rendering with limited thermal accumulation throughout the tested operating period.

Beyond canonical square-wave and sinusoidal references, the actuator was evaluated using a structured interaction-like profile and torque trajectories generated from an intracerebral-infusion biomechanics model. The latter experiment establishes a complete signal chain from the infusion flow rate and tissue permeability to the simulated pressure response, torque reference, and physically rendered resistive torque. The actuator reproduced the main temporal features of the model-derived trajectories across low-, medium-, and high-amplitude feedback conditions.

Overall, the representative studies in Table S2 have established important capabilities in compact torque generation, multidirectional actuation, wearable interaction, medical haptics, and low-power rendering. The present work further integrates magnetic self-sealing, PWM-dependent open-loop characterization, hysteresis-aware closed-loop torque regulation, sustained thermal stability, and biomechanics-model-informed rendering within a single rotary haptic actuator. This integrated hardware–drive–control framework advances MR-based haptic actuation from individual device-function demonstrations toward a more systematic and practically implementable torque-rendering platform.

## Supplementary Material S3: Detailed Mason-Number Assessment of the Magnetic Sealing Region

The main manuscript uses the Poiseuille–Couette framework of Liang *et al.* [12] to assess the flow regime within the permanent-magnet-assisted sealing region of the proposed MR haptic actuator. The sealing gap is locally approximated as a thin planar slot subjected to a pressure-driven load and motion of one bounding surface. For the nominal geometry,

$$\frac{\delta}{R} = \frac{0.1\ \mathrm{mm}}{3\ \mathrm{mm}} \approx 0.033 \ll 1, \tag{S1}$$

supporting the local thin-gap approximation.

Within the Bingham-fluid representation adopted by Liang *et al.*, the relative effects of the pressure gradient, moving-wall shear, and field-dependent yield stress are described by

$$\begin{aligned} Mn(p) &= \frac{\delta}{\tau_y}\left|\frac{dp}{dx}\right|, \\ Mn(\Omega) &= \frac{\tau_y \delta}{\mu R\,|\Omega|}, \end{aligned} \tag{S2}$$

where $\delta$ is the sealing-gap width, $\tau_y$ is the field-dependent yield stress, $\mu$ is the effective post-yield viscosity, $R$ is the characteristic radius of the moving boundary, and $\Omega$ is its angular velocity. The transition boundaries from the one-region to the two-region mode and from the two-region to the three-region mode are [12]

$$\begin{aligned} Mn(p)_{R1} &= \frac{2}{Mn(\Omega)}, \\ Mn(p)_{R2} &= 2 + \frac{1}{Mn(\Omega)} + \sqrt{\frac{1}{Mn(\Omega)^2} + \frac{4}{Mn(\Omega)}}. \end{aligned} \tag{S3}$$

### A. Gravity-Induced Pressure Load

A conservative gravity-loaded condition was considered with the actuator oriented vertically. The full weight of the contained MRF was assumed to act on the lower annular sealing interface. The nominal contained MRF volume and density were

$$V_{\mathrm{MRF}} = 640.326\ \mathrm{mm}^3 = 6.40326 \times 10^{-7}\ \mathrm{m}^3, \qquad \rho_{\mathrm{MRF}} = 3.0 \times 10^3\ \mathrm{kg/m}^3. \tag{S4}$$

The corresponding MRF mass was

$$m_{\mathrm{MRF}} = \rho_{\mathrm{MRF}} V_{\mathrm{MRF}} = 1.921 \times 10^{-3}\ \mathrm{kg}. \tag{S5}$$

The effective annular area was determined from the shaft radius $r_i = 3.0$ mm and the outer sealing radius $r_o = 3.1$ mm:

$$\begin{aligned} A_s &= \pi \left(r_o^2 - r_i^2\right) \\ &= \pi \left[\left(3.1 \times 10^{-3}\right)^2 - \left(3.0 \times 10^{-3}\right)^2\right] \\ &\approx 1.916 \times 10^{-6}\ \mathrm{m}^2. \end{aligned} \tag{S6}$$

The resulting equivalent gravity-induced pressure difference was therefore

$$\begin{aligned} \Delta P_g &= \frac{m_{\mathrm{MRF}} g}{A_s} \\ &= \frac{\left(1.921 \times 10^{-3}\right)(9.81)}{1.916 \times 10^{-6}} \\ &\approx 9.83 \times 10^3\ \mathrm{Pa} \approx 9.8\ \mathrm{kPa}. \end{aligned} \tag{S7}$$

The characteristic sealing length was taken as the effective leakage-path length across the magnetically active zone, $L = 2$ mm. The corresponding pressure-gradient magnitude was

$$\left|\frac{dp}{dx}\right| \simeq \frac{\Delta P_g}{L} = \frac{9.8 \times 10^3}{2 \times 10^{-3}} \approx 4.9 \times 10^6\ \mathrm{Pa/m}. \tag{S8}$$

### B. Rotary and Material Parameters

The representative mechanical rotation frequency used in the sealing assessment was

$$f_{\mathrm{rot}} = 2\ \mathrm{Hz}, \qquad \Omega = 2\pi f_{\mathrm{rot}} = 4\pi\ \mathrm{rad/s}. \tag{S9}$$

A representative sealing-gap flux density of approximately 0.2 T was extracted from the magnetostatic field solution. The corresponding field-dependent yield stress was taken as

$$\tau_y = 7.83\ \mathrm{kPa} \tag{S10}$$

based on the material data in [13]. The effective post-yield viscosity used in the Bingham-fluid approximation was

$$\mu = 0.112\ \mathrm{Pa\,s}. \tag{S11}$$

Using these parameters, the gap-based Reynolds number is

$$\begin{aligned} \mathrm{Re}_\delta &= \frac{\rho_{\mathrm{MRF}} R \left|\Omega\right| \delta}{\mu} \\ &= \frac{\left(3.0 \times 10^3\right)\left(3 \times 10^{-3}\right)(4\pi)\left(0.1 \times 10^{-3}\right)}{0.112} \\ &\approx 0.10 < 1. \end{aligned} \tag{S12}$$

Together with the small gap-to-radius ratio in Eq. (S1), this result supports the low-inertia, locally planar Poiseuille–Couette approximation used for the regime assessment.

*C. Mason-Number Evaluation*

Substitution into Eq. (S2) gives

$$\begin{aligned} Mn(p) &= \frac{0.1\times10^{-3}}{7.83\times10^{3}}\left(4.9\times10^{6}\right)\approx 0.0626,\\ Mn(\Omega) &= \frac{\left(7.83\times10^{3}\right)\left(0.1\times10^{-3}\right)}{(0.112)\left(3\times10^{-3}\right)(4\pi)}\approx 185.4. \end{aligned} \tag{S13}$$

The corresponding transition values obtained from Eq. (S3) are

$$Mn(p)_{R1}\approx 0.0108, \qquad Mn(p)_{R2}\approx 2.152. \tag{S14}$$

Thus,

$$Mn(p)_{R1} < Mn(p) < Mn(p)_{R2}, \tag{S15}$$

placing the sealing flow in the *two-region mode*. The nominal margins relative to the two transition boundaries are

$$\frac{Mn(p)}{Mn(p)_{R1}}\approx 5.80, \qquad \frac{Mn(p)}{Mn(p)_{R2}}\approx 0.029. \tag{S16}$$

In the two-region mode, the model predicts a plug region attached to the magnet-side wall of the sealing gap, while the remaining portion of the gap is yielded and accommodates the relative wall motion [12]. The wall-attached plug region therefore provides additional resistance to pressure-driven MRF migration during rotary motion.

*D. Parameter-Sensitivity Assessment*

The sensitivity of the regime classification to uncertainty in $\tau_y$, $\delta$, and $L$ was first assessed by independently varying each parameter by $\pm 20\%$, while holding the remaining quantities at their nominal values. The lower-bound margin can be written as

$$\begin{aligned} \frac{Mn(p)}{Mn(p)_{R1}} &= \frac{Mn(p)Mn(\Omega)}{2}\\ &= \frac{\delta^2}{2\mu R\,|\Omega|}\left|\frac{dp}{dx}\right|\\ &= \frac{\delta^2\Delta P_g}{2\mu R\,|\Omega|\,L}. \end{aligned} \tag{S17}$$

The yield stress therefore cancels from the lower-bound margin, although it continues to affect the individual Mason numbers and the upper transition boundary.

For the one-at-a-time variations, the smallest lower-bound margin occurred when the sealing-gap width was reduced by 20%:

$$\left.\frac{Mn(p)}{Mn(p)_{R1}}\right|_{\delta=0.8\delta_0} = 5.80\,(0.8)^2\approx 3.71>1. \tag{S18}$$

An additional combined-corner assessment was performed by evaluating all combinations in which $\tau_y$, $\delta$, and $L$ simultaneously took their $\pm 20\%$ limits. The smallest lower-bound margin occurred for

$$\delta = 0.8\delta_0, \qquad L = 1.2L_0, \tag{S19}$$

and was

$$\left.\frac{Mn(p)}{Mn(p)_{R1}}\right|_{\text{combined}} = 5.80\frac{(0.8)^2}{1.2}\approx 3.09>1. \tag{S20}$$

Across the same combined uncertainty cases, the largest upper-bound ratio remained

$$\max\left[\frac{Mn(p)}{Mn(p)_{R2}}\right]\approx 0.055<1. \tag{S21}$$

Consequently, none of the one-at-a-time or combined $\pm 20\%$ parameter cases changed the predicted two-region classification.

The present Mason-number calculation is used as a flow-regime assessment rather than as a direct prediction of leakage rate. No visible MRF leakage was observed during the characterization and controller-tuning experiments reported in the main manuscript, providing preliminary experimental support for magnetic fluid confinement under the tested conditions. Quantitative leakage-rate measurement, cyclic sealing endurance, and long-duration orientation-dependent testing remain subjects for future investigation.



\end{document}